\documentclass{article}
\pdfoutput=1
\usepackage[utf8]{inputenc} 
\usepackage[T1]{fontenc}    
\usepackage{hyperref}       
\usepackage{url}            
\usepackage{booktabs}       
\usepackage{amsfonts}       
\usepackage{nicefrac}       
\usepackage{microtype}      
\usepackage{graphicx}
\usepackage{kotex}
\usepackage{amsmath}

\usepackage[numbers]{natbib}
\usepackage[final]{nips_2017}
\usepackage{gensymb}
\title{
Unsupervised Visual Attribute Transfer with Reconfigurable Generative Adversarial Networks
}

\author{
Taeksoo Kim, Byoungjip Kim, Moonsu Cha, Jiwon Kim \\
SK T-Brain \\
\texttt{\{jazzsaxmafia,bjkim,ckanstnzja,jk\}@sktbrain.com} \\
}

\usepackage{xcolor}

\begin{document}

\maketitle
\begin{abstract}

Learning to transfer visual attributes requires supervision dataset. Corresponding images with varying attribute values with the same identity are required for learning the transfer function. This largely limits their applications, because capturing them is often a difficult task. To address the issue, we propose an unsupervised method to learn to transfer visual attribute. The proposed method can learn the transfer function without any corresponding images. Inspecting visualization results from various unsupervised attribute transfer tasks, we verify the effectiveness of the proposed method. 

\end{abstract}
\section{Introduction}
What would you look like if you grow bang hair, dye your hair blond, or change your gender? Recent image-to-image translation research \cite{isola2017im2im,kim2017discogan,zhu17cycle} can be the solutions for these demands. By learning a fixed mapping function from a source domain to a target domain, it can make one have bang hair or change her hair color into a specific color. More recently, image-to-image translation with unpaired dataset \cite{kim2017discogan,zhu17cycle} has been researched, getting rid of the necessity of tedious data collection. 

However, what if you do not just want a bang hair, but exactly the one that Uma Thurman had in the movie Pulp Fiction? What about having the exact smile of Mona Risa? This is the part where image-to-image translation cannot help due to its nature of treating the whole domain instead of each instance. 

On the other hand, researches on image stylization \cite{colortransfer, styletransfer, texturetransfer, lin2015iccv, dos2016nips}, that address instance-level visual attribute transfer have drawn  significant amount of attention recently. Despite the remarkable results of these works, most of these works are limited to holistic attribute transfer. For example, they transfer texture, color or style that cover the whole image but not specific part of an image like smile or hair color.



We therefore seek a visual attribute transfer mechanism that is able to transfer certain part of an image instance-level, without paired dataset. First, we formally define \textit{attributes} as high-level features of a dataset that are independent to each other, and can be disentangled in well-defined feature space (hair color, bangs). We further denote \textit{attribute values} as set of possible values for each attribute that can be easily labeled (black/blond hair, with/without bangs), and \textit{domain} as the set of data that have labeled as certain attribute value (images of people with bangs). Each element in an attribute domain is unique in its details (every bang hair is different) so that instance level annotation is infeasible. 

Since we have only access to domain-level labels, we exploit domain transfer methods previously proposed by \cite{kim2017discogan,zhu17cycle} along with an additional objective that enforces instance-level transfer. After transferring the target attribute of a reference image to the source image, in order to encourage the transferred result image to belong to the target attribute domain, we use generative adversarial network (GAN) \cite{goodfellow2014generative} so that the result image is indistinguishable from the images in target domain. In order to ensure the remaining non-target attributes are kept intact, we introduce \textit{back-transfer} objective inspired by \cite{kim2017discogan,zhu17cycle}. The target attribute of the source image is transferred back to the result image, and it is forced to be the same as the original source image. Furthermore, in order to force instance-level transfer, we introduce \textit{attribute-consistent} objective: we transfer the target attribute of the result image back to the reference image, and encourage this to be the same as the original reference image. This objective pushes the result image to have the exact instance-level attribute details of the reference image.

Our proposed framework has three distinctive features: 1) instead of changing an attribute of the source image using a fixed mapping function like image-to-image translation does, it can transfer the instance-level attribute. Our method also works well for domain-level transfer, which corresponds to translation, as shown in various experiments. 2) It does not require paired dataset, nor densely annotated attribute information. By only using domain-level labeled data, our proposed model successfully transfers the details of a reference image to the source image. 3) It can be used to change multiple attributes of the source images using a single model, while image-to-image translation based methods require training new model for each.

This paper is structured as follows. We summarize the related works in Section 2 and discuss relations and differences. In Section 3, we present the architecture and training method of the proposed model. Our experiment results are detailed in Section 4. We conclude the paper in Section 5.
\section{Related Work}

\subsection{Image-to-Image Translation}
In computer vision and graphics, the concept of image-to-image translation is used to deal with a set of related problems. The goal of image-to-image translation is to learn the mapping from an input image of a source domain $A$ to an output image of a target domain $B$. Isola et al. \cite{isola2017im2im} proposed a framework called pix2pix, which used generative adversarial networks for image-to-image translation. This approach can be categorized as paired image-to-image translation, since the model requires paired image data for training. However, obtaining paired training data is usually expensive. To address this issue, unpaired image-to-image translation \cite{kim2017discogan,zhu17cycle} has been proposed. This approach enables to learn the mapping between domains without paired input-output examples. However, to generate an output image where multiple effects are applied (e.g., a woman with black hair $\rightarrow$ a man with brown hair), this approach requires to build many models for each simple translation. What differentiates our model is that single model enables such multiplex translations in a shared and reconfigurable manner. With similar aspects, there are research efforts classified as conditional image generation. Xinchen Yan et al. \cite{yan2016att2im} proposed a conditional image generation model called Attribute2Image by using Variational Auto-encoders (VAEs) \cite{kingma2014vae}. However, this approach requires training image data that are densely labeled by attributes. In contrast, our model does not require such densely-labeled training data. 

\subsection{Generative Models}
Generative models \cite{goodfellow2014generative, kingma2014vae, kulkani2015dcign, larsen2016vaegan, mirza2014conditional} are trained with large amount of data to generate similar data to the training data. More formally, the objective of generative models is to minimize the difference between data distribution \textit{p\textsubscript{data}(x)} and generation distribution $p_{g}(x)$. In this problem setting, there are some promising generative models including Variational Autoencoders (VAEs) \cite{kingma2014vae} and Generative Adversarial Nets (GANs) \cite{goodfellow2014generative}. 

VAEs approach the generation problem by using the framework of probabilistic graphical models, and the objective is to maximize a lower bound on the log likelihood of the generation distribution $p_{\theta}(x|z)$. 

\begin{equation}
\mathbb{E}_{q_{\phi}(z|x)}[-\log(p_{\theta}(x|z))] + KL(q_{\phi}(z|x) || p(z))
\end{equation}

GANs approach the generation problem by using the adversarial learning framework that is considered as a game between two separate networks: a generator and a discriminator. The discriminator examines whether samples are real or fake. The generator tries to confuse the discriminator by gradually learning data distribution $p_{data}(x)$. Since GANs enable to learn a loss function instead of specifying it explicitly, they can usually generate very realistic data. However, it is known that training GANs is rather hard due to the dynamics of the adversarial training.

\begin{equation}
\min_{G} \max_{D} V(D,G) = \mathbb{E}_{x \sim p_{data}(x)}[\log D(x)] + \mathbb{E}_{z \sim p_{z}(z)}[\log(1-D(G(z)))]
\end{equation}

\subsection{Disentangled Representations}
Disentangled representations \cite{bengio2013deep,bengio2013disrepre} allow conditional generative models that can generate sample data by changing corresponding factor codes. More formally, conditional generative models are described by $p_{g}(x|z,c)$ where $z$ is a latent code and $c$ is a disentangled factor code. Very recently, researchers show a possibility that disentangled representations can be learned with GANs. For example, Xi Chen et al. \cite{chen2016infogan} proposed InfoGAN, an extension of GAN that can learn disentangled representations for images. To learn disentangled representations (e.g., the width of digits), InfoGAN adds new objective that maximizes the mutual information between a factored code and the observation. Michael Mathieu et al. \cite{mathieu2016cvaegan} proposed a conditional generative model based on conditional VAE. They addressed degeneration problem by adding the GAN objective into the VAE objective.
\section{Formulation}

\begin{figure}[t]
  \begin{center}
  \centerline{\includegraphics[width=\textwidth] {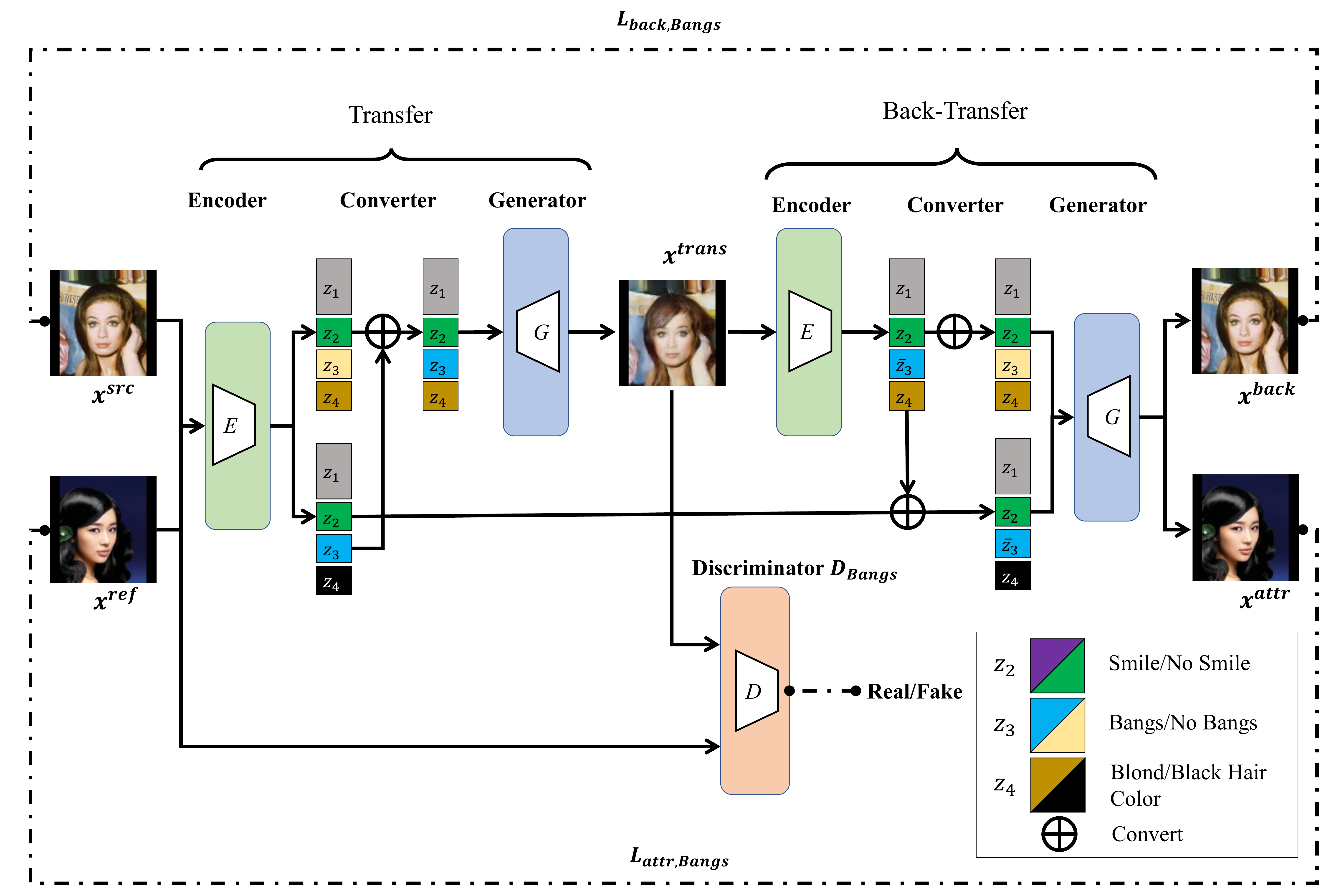}}
  \caption{The network architecture of proposed model. In this figure, we show an example of bang attribute transfer. In \textit{transfer} phase, the target attribute of the reference image is transferred into the input image, and the result $x^{trans}$ is encouraged to be indistinguishable from real images with bang hair. In \textit{back-transfer} phase,  the target attribute of the result image is transferred back into the input image.  In order to enforce $x^{trans}$ to have the details of the reference image, the target attribute of $x^{trans}$ is transferred into the reference image, which is enforced to be the same as the original reference image.}
  \label{fig:archi}
  \end{center}
\end{figure}

The basic building blocks of our framework are an encoder $E$, a generator $G$ and discriminator $D_{v}$ for each attribute value $v$ of total $m$ attribute values. Our model can perform transfer for multiple attributes at the expense of one encoder and one generator, along with as many discriminators as the total number of attribute values. $E$ takes an image $x$ as input and encodes it into attribute features $z$ as $z=E(x)$. We assume the attributes can be disentangled in well-defined feature space, so that $z$ can be expressed as a tuple of multiple \textit{slots}: $z=\{{z_{1}, z_{2},...z_{n}}\}$, where $n$ is the number of attributes of interest and $z_1$ is reserved as sample-specific attribute. Note the number of attribute $n$ is different from the number of attribute values $m$. $G$ takes the encoded attribute feature and generate an image: $y=G(z_{1}, z_{2},...z_{n})$, where $y$ indicates generated image.

In order to transfer target attribute from a reference image to the source image, both images $x^{src} \sim p_{data}(x)$ and $x^{ref} \sim p_{v}(x)$ are first encoded: $z^{src}=E(x^{src})$ and $z^{ref}=E(x^{ref})$, where $p_{data}(x)$ is the data distribution of the entire dataset and $p_{v}(x)$ denotes the distribution of the domain of attribute value $v$. We omit the notation for the attribute index for simplicity. The target slot of $z^{src}$ is then replaced by that of $z^{ref}$, then $G$ takes this attribute features to generate the result image, $x^{trans}=G(z^{src}_{1},z^{src}_{2},...,z^{ref}_{tar},...,z^{src}_{n})$


In order for this image to have the target attribute from the reference image, we impose three constraints. First of all, it has to belong to the same target attribute domain as the reference image. Also, it needs to keep other attributes intact after transference. Finally, it should have the exact details of the reference image's target attribute. To encourage these three constraints to be satisfied, we impose objective for each constraint. We explain each of objective in detail below.

\paragraph{Transfer}
Transferring target attribute so that the result image to belong to the corresponding domain can be implemented using GAN, where we denote this objective as \textit{transfer objective}. Consider the target attribute corresponds to hair color, and the reference image has black hair as attribute value, then we enforce $x^{trans}$ to be indistinguishable from images in black hair domain. Transfer objective for the target attribute value is expressed as follows: 
    \begin{align}
    	L_{trans,v} &= \mathbb{E}_{x \sim p_{data}(x)}[\log D_{v}(x^{trans})] \label{eq:obj_1}
    \end{align}
Also, the discrimination objective of GAN is:
    \begin{align}
    	L_{{dis},v} &= \mathbb{E}_{x \sim p_{data}(x)}[\log(1-D_{v}(x^{trans}))] + \mathbb{E}_{x^{ref} \sim p_{v}(x)}[\log D_{v}(x^{ref})]\label{eq:obj_12}.
    \end{align}

\paragraph{Back-transfer}
To encourage all of the remaining non-target attributes to be kept intact, we introduce \textit{back-transfer objective}, inspired by \textit{cycle-consistency objective} \cite{kim2017discogan,zhu17cycle}. Transferred image $x^{trans}$ is again encoded as $z^{trans}=E(x^{trans})$, then the original target slot of $z^{src}$ replaces that of $z^{trans}$ as $z^{back}=\{z^{trans}_{1},z^{trans}_{2},...,z^{src}_{tar},...,z^{trans}_{n}\}$. We impose the generated image from this feature to be the same as the original source image: $x^{back} = G(z^{back}) \approx x^{src}$.
    \begin{align}
    	L_{back,v} &= \mathbb{E}_{x^{src} \sim p_{data}(x)}[dist(x^{src}, x^{back})]  \label{eq:obj_3}
    \end{align}
This objective enforces all the non-target attributes of transferred image to be the same as those of the source image.
\paragraph{Attribute consistency}
Training a model with transfer and back-transfer objective ensures the transferred image to have the target attribute value, while the remaining non-target attributes are kept intact. However, these objectives do not ensure the transferred image to have the exact attribute details of the reference image. For example, it can have any type of bang, as long as it is indistinguishable from the images in bang hair domain. We therefore introduce \textit{attribute-consistency} objective that encourages the transference of the details: $x^{attr} = G(z^{ref}_{1},z^{ref}_{2},...,z^{trans}_{tar},...,z^{ref}_{n}) \approx x^{ref}$. This can be expressed formally as,
    \begin{align}
    	L_{attr,v} &= \mathbb{E}_{x^{ref} \sim p_{v}(x)}[dist(x^{attr}, x^{trans})]  \label{eq:obj_4}
    \end{align}

The distance $dist$ can be chosen among any metrics including $L1$, $L2$ or Huber.  In order this objective to be satisfied, target attribute feature of the transferred image $z^{trans}_{tar}$ need to encode the details of the target attribute of the reference image.

\paragraph{Full objective}
The full transfer objective for attribute value $d$ is:
    \begin{align}
    	L_{gen,v} &=  \lambda_{1}L_{trans,v} +\lambda_{2}L_{back,v} +\lambda_{3}L_{attr,v} \label{eq:obj_5}
    \end{align}
 where $\lambda_{1}$, $\lambda_{2}$, $\lambda_{3}$ are importance weights of each objective.
During training, for each attribute value $v$, the parameters of $E$ and $G$ are updated using $L_{gen,v}$, and the parameters of $D_v$ are updated using $L_{dis,v}$. More specifically, each iteration of training is composed of $m$ (total number of attribute values). In each step, reference images are sampled from $v$'s domain, then $E$, $G$ and $D_v$ are updated using $L_{gen,v}$ and $L_{dis,v}$. This procedure is conducted for all $m$ attribute values every iteration. The architecture of our model and training process for a specific $v=Bangs$ is depicted in Fig. \ref{fig:archi}.

\section{Experiments}

In this section, we evaluate the performance of our proposed method. First, we show the results of instance-level attribute transfer for human face images. Specific facial attributes such as hair color or smile of reference images are transfered to input images. We also conduct domain-level attribute transfer experiments on several tasks: facial attribute, angle/object, and fashion garments . Domain-level attribute transfer changes a specific attribute values using fixed function. It is done using the average attribute vector of the target attribute value instead of the attribute vector of a specific reference image $z^{ref}_{tar}$. This average attribute vector is obtained by averaging the attribute vectors of all images in the domain of target attribute value. We then show the experiment results of ``multiplex'' case where the multiple domain-level attributes are transferred simultaneously.

When training models for domain-level transfer tasks, mini-batch average of attribute vectors were used instead of full batch average in each iteration. We also set $\lambda_{3}=0$ in (\ref{eq:obj_5}) since we are no longer treating single reference image in domain-level task.

\subsection{Instance-level Attribute Transfer}
\begin{figure}[h!]
  \begin{center}
  \centerline{\includegraphics[width=0.8\textwidth]{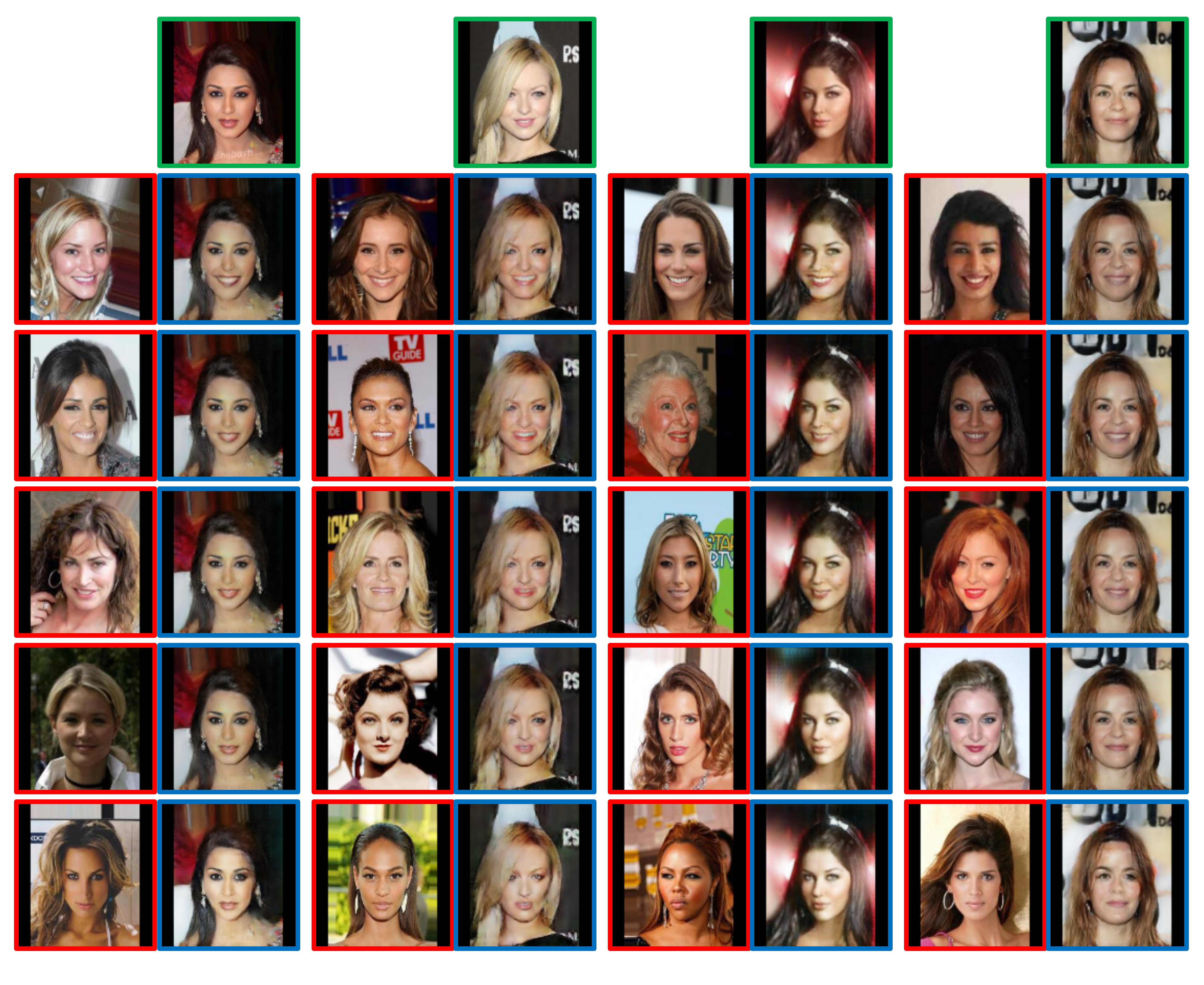}}
  \caption{Results of smiling attribute transfer. Images with green, red and blue bounding boxes indicate source, reference and transferred results, respectfully. }
  \label{fig:attr_1}
  \end{center}
\end{figure}

\begin{figure}[h!]
  \begin{center}
  \centerline{\includegraphics[width=0.8\textwidth]{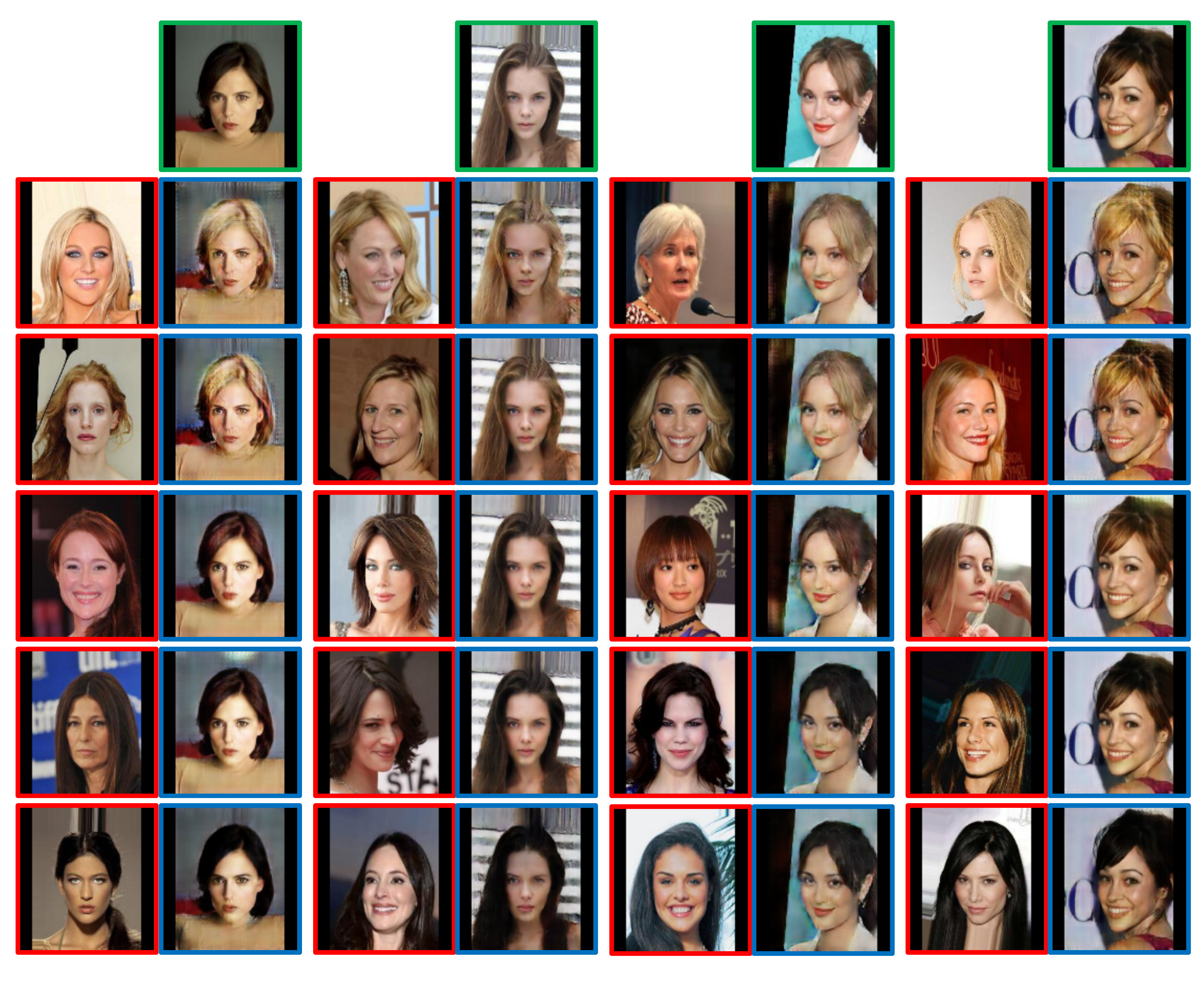}}
  \caption{Results of hair color attribute transfer. Images with green, red and blue bounding boxes indicate source, reference and transferred results, respectfully.}
  \label{fig:attr_2}
  \end{center}
\end{figure}

\begin{figure}[h!]
  \begin{center}
  \centerline{\includegraphics[width=0.8\textwidth]{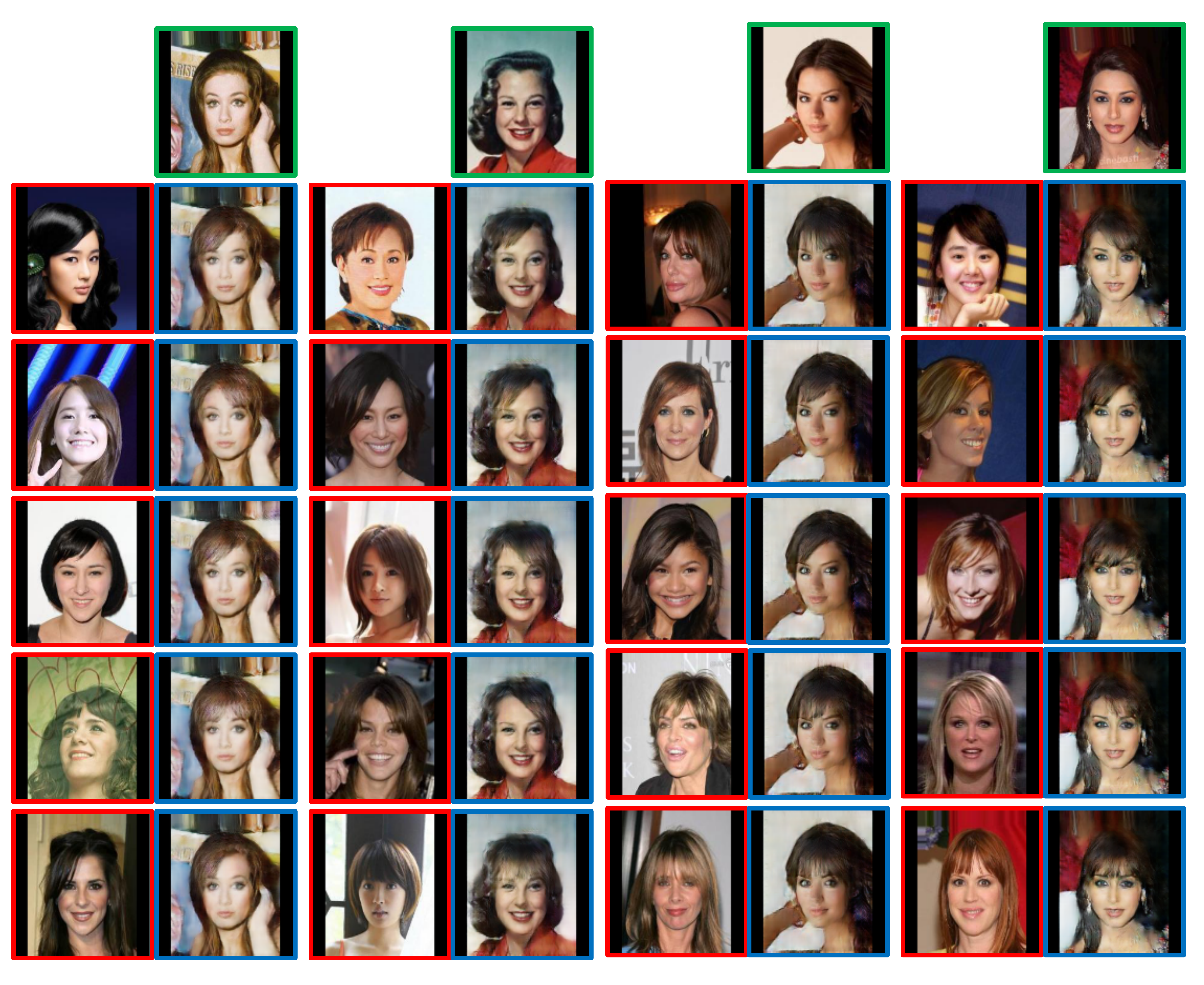}}
  \caption{Results of bang hair attribute transfer. Images with green, red and blue bounding boxes indicate source, reference and transferred results, respectfully.}
  \label{fig:attr_3}
  \end{center}
\end{figure}

We use the CelebA dataset \cite{liu2015faceattributes} that consists of images of human face, along with the annotations of their attributes. We used three attributes: hair color, bang and smiling. For each attribute we used two attribute values: blonde/black for hair color, with/without bang for bang and with/without smile for smiling. The results for visual attribute transfer is shown in Fig. \ref{fig:attr_1}, \ref{fig:attr_2} and \ref{fig:attr_3}. Note that these are results of a single model rather than three different models. In each example, the details of the reference image is transferred into the input image.

\subsection{Domain-level Attribute Transfer}

\subsubsection{Face}
  We use previously used CelebA dataset for domain-level attribute transfer experiments. We use the following visual attributes and their values: hair colors (blond/black/brown), bang (with/without bang) and smiling(with/without smile).

\begin{figure}[h!]
  \begin{center}
  \centerline{\includegraphics[width=\textwidth]{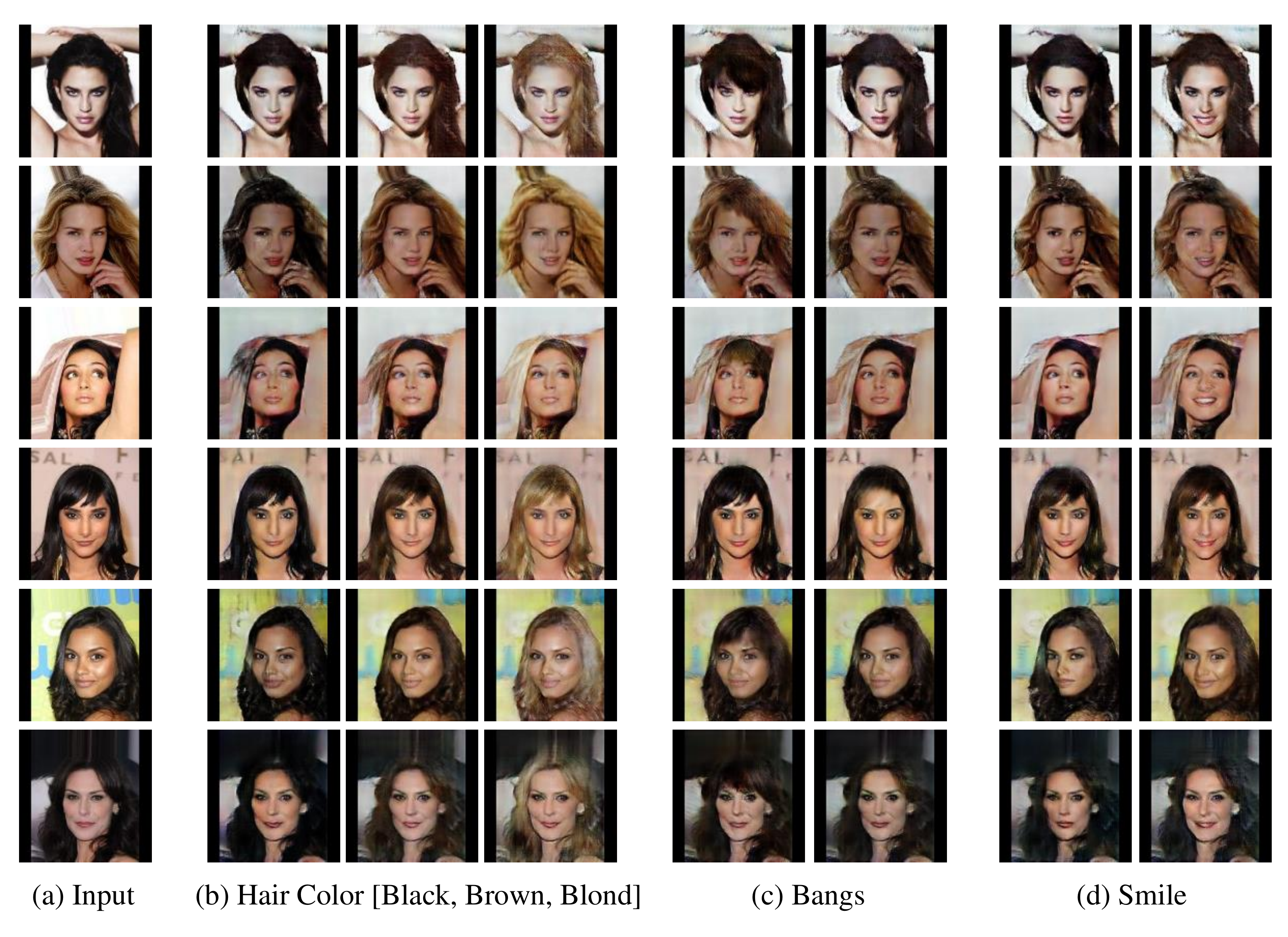}}
  \caption{Domain-level attribute transfer results for CelebA dataset. A specific attribute of input images (a) are transferred into the target values. Unlike instace-level transfer, where a target attribute vector of a reference image is used, an average vector of the attribute value is used for domain-level transfer. (b) Hair color transfer into black, brown and blonde, (c) bang attribute transfer, (d) smile attribute transfer.}
  \label{fig:exp_face_1}
  \end{center}
\end{figure}

\begin{figure}[h!]
  \begin{center}
  \centerline{\includegraphics[width=\textwidth]{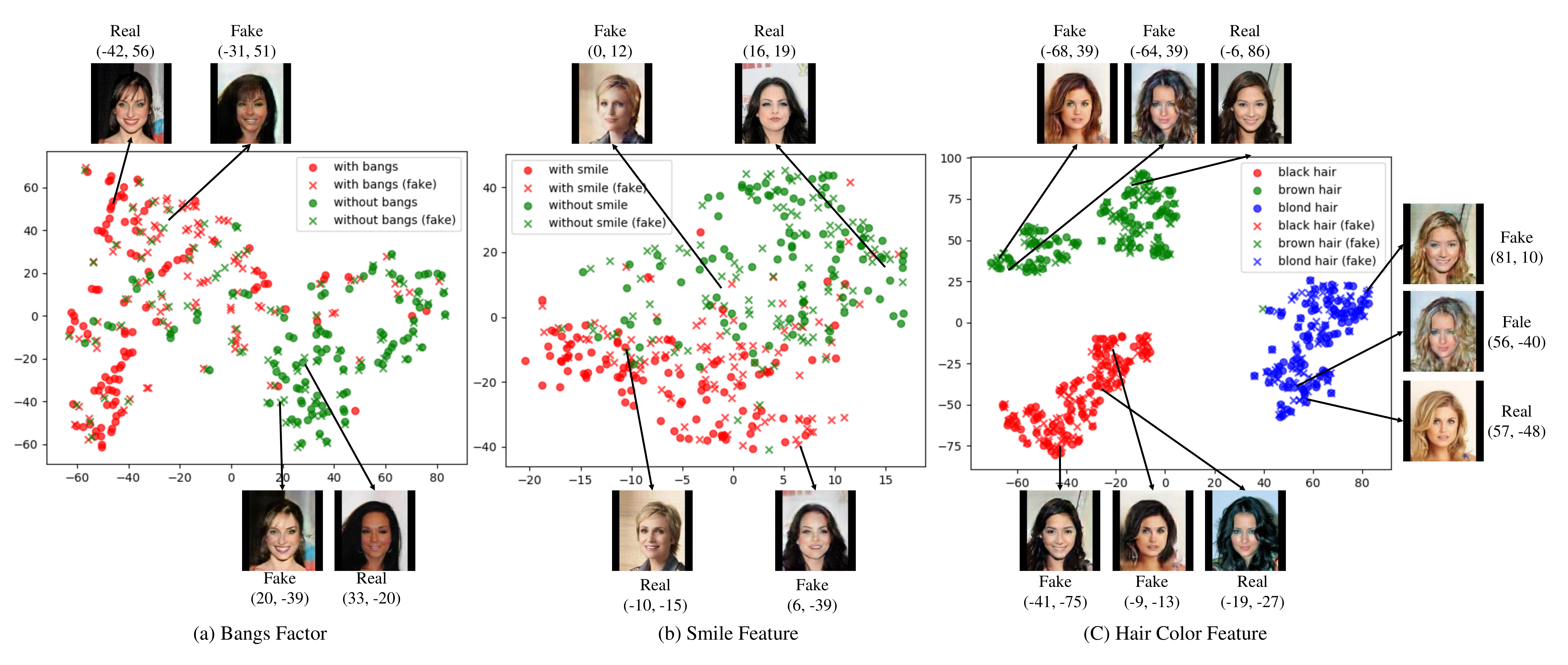}}
  \caption{t-SNE visualization of attribute vectors of the face dataset. (a) shows the distribution of the "bang hair" vectors. Red circles are real "bang hair" images, and red crosses are generated "bang hair" images. (b) shows the distribution of the "smile" representations. (c) shows the distribution of the "hair color" representations. For every plot, colored x's indicate encoded attribute vectors obtained from transferred images.}
  \label{fig:tsne}
  \end{center}
\end{figure}

Our results shown in Fig. \ref{fig:exp_face_1} illustrate that the proposed method successfully transfers desired visual image attributes while maintaining all other attributes. Also, as shown in the t-SNE plot of Fig. \ref{fig:tsne}, the encoder learns to separate different attribute values, and the distribution of attributes from the result images closely match that from real images.

\subsubsection{Angle/Object}
\begin{figure}[t]
  \begin{center}
  \centerline{\includegraphics[width=0.8\textwidth]{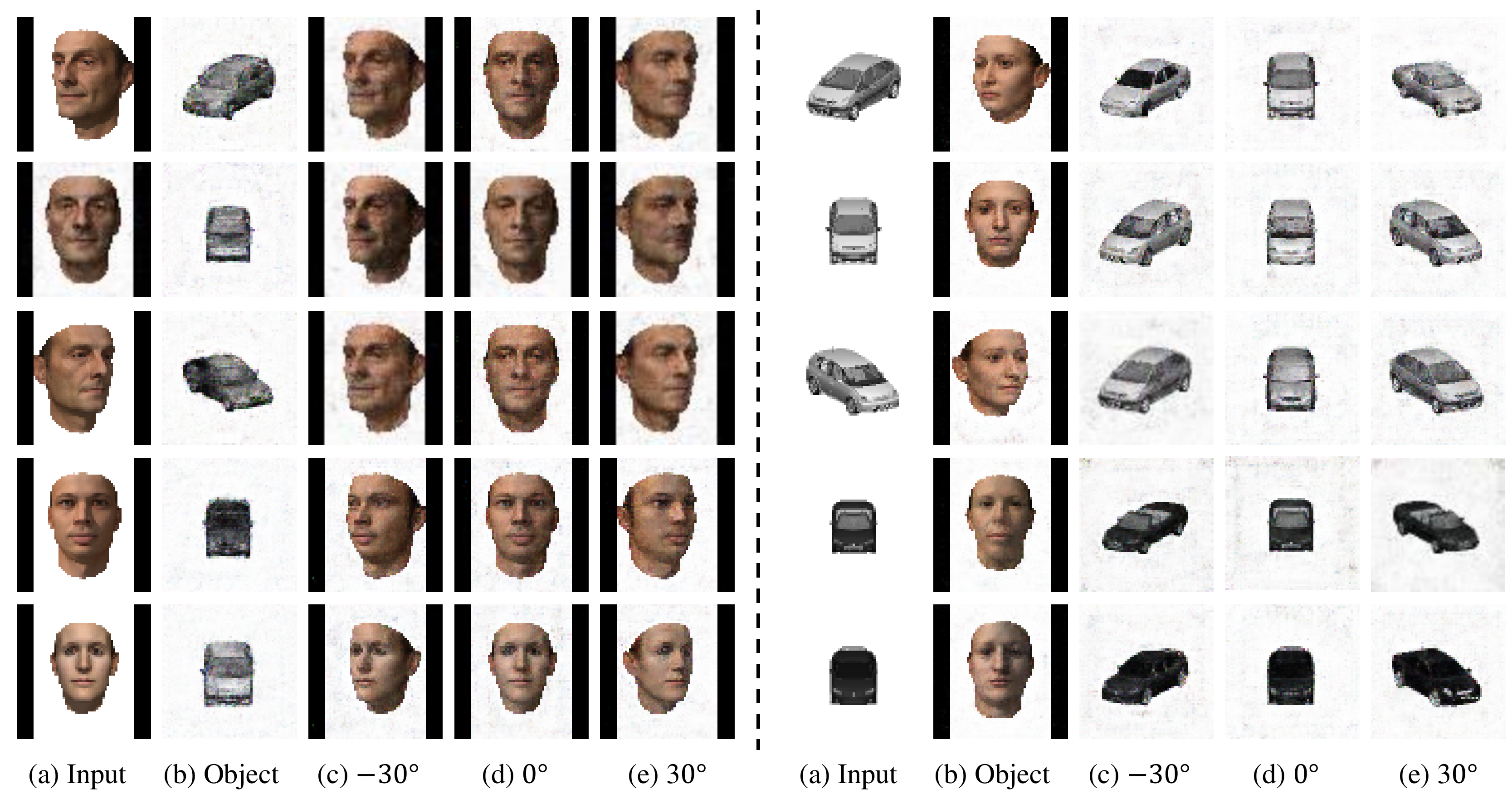}}
  \caption{Domain-level transfer of object type and azimuth angle attributes. The left part of the figure shows the case of face inputs. Column (b) shows the results of transferring the input image's object type attribute from face to car. As shown in the figure, while maintain the azimuth angle attribute, the object type attribute is successfully changed. Columns (c), (d) and (e) show the results of transferring the azimuth angle attribute to -30, 0, and 30 degrees respectively. The right half of the figure  shows the case of car inputs. Similar to the case of face inputs, the results show successful attribute transfer.}
  \label{fig:exp_face_car_1}
  \end{center}
\end{figure}

We consider a case where the visual attributes of two image domains differ greatly. To investigate whether our model can cope with this, we combined two rendered 3D image datasets: 3D Car \cite{fidler2012car} and 3D Face \cite{bfm09}. The combined dataset consists of images of human faces and cars that rotate with respect to azimuth angle. We use the object type and angle as visual attributes. The object type attribute can take on value of car or face, and the angle attribute takes on either of the three angles: -30, 0, 30 degrees. The transfer results are shown in the figure \ref{fig:exp_face_car_1}.

\begin{figure}[h!]
  \begin{center}
  \centerline{\includegraphics[width=0.8\textwidth]{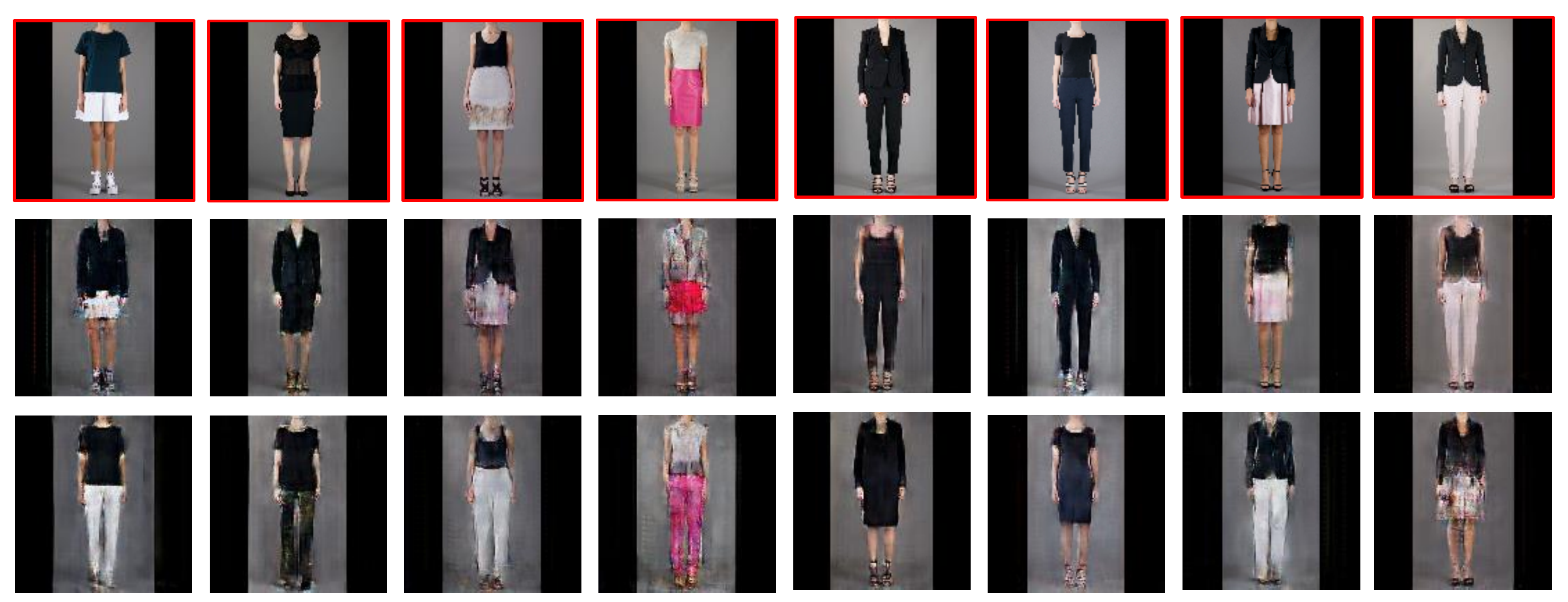}}
  \caption{Domain-level transfer of fashion garment attributes. Images on the first row are the inputs. Result images where the bottoms are changed (trousers to skirt, skirt to trousers) is shown in the second row. The last row indicate generated images where the tops are changed (jacket to top, top to jacket)}
  \label{fig:exp_fashion_1}
  \end{center}
\end{figure}


\subsubsection{Fashion}

  In order to further investigate the attribute transfer ability of our model, we conduct a fashion item transfer. In this experiment, our model receives an input image of a person wearing specific tops (with/without jacket) and bottoms (trousers/skirt). We show that our model can change one of top and bottom garments, while preserving other attributes intact in Fig. \ref{fig:exp_fashion_1}.

\begin{figure}[h!]
  \begin{center}  \centerline{\includegraphics[width=0.8\textwidth]{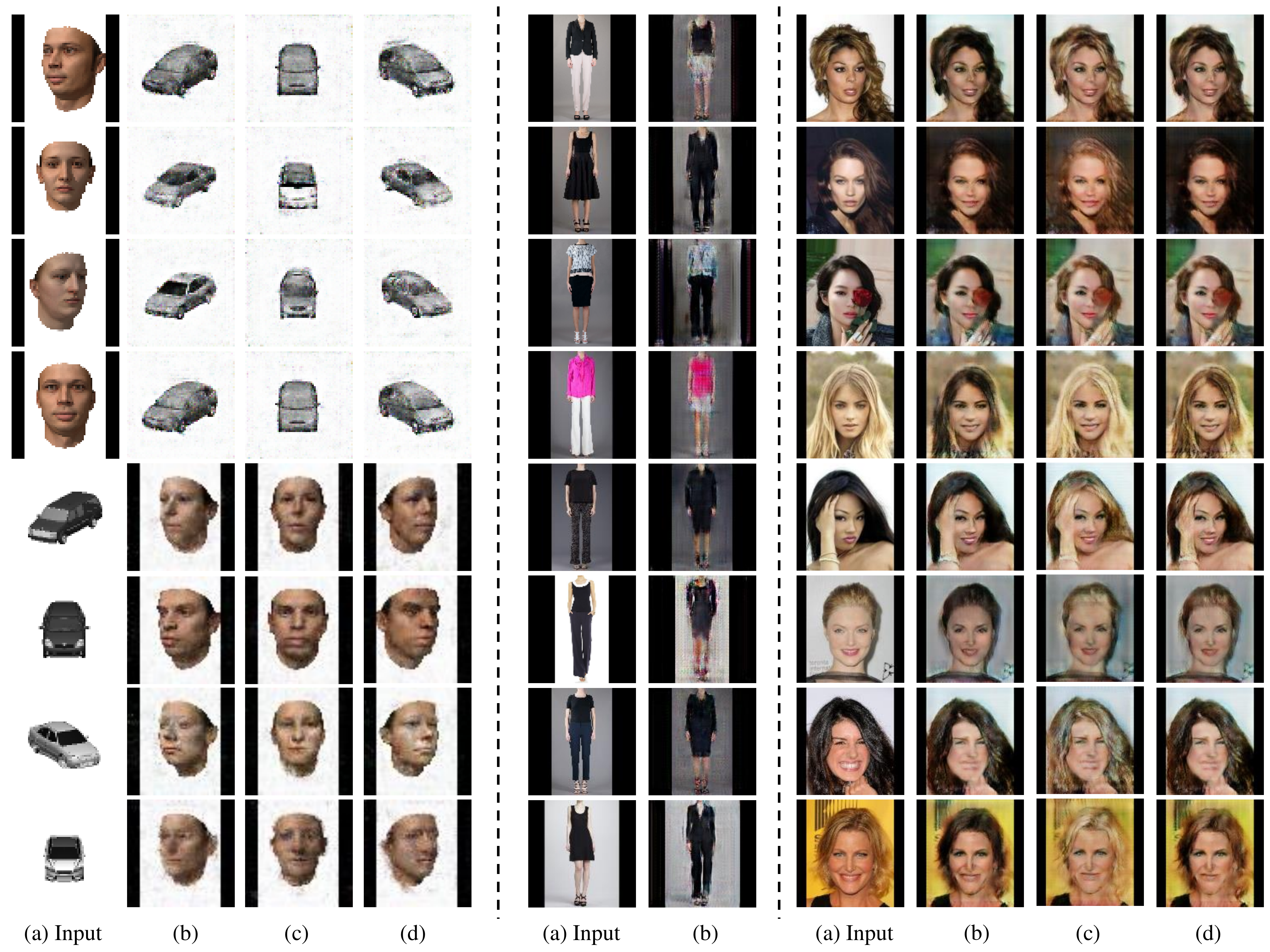}}
  \caption{Multiplex image attribute transfer. The left columns shows the results based on multiplex object or angle attributes. Faces and cars are inputs, and different angles of different objects are targets. The results are as follows: (b) -30$\degree$, object type attribute; (c) 0$\degree$, object type attribute; (d) 30$\degree$, object type attribute transfer.
The middle columns are results of multiplex fashion attribute transfer. Target images have opposite attributes of input images, and it transfers the input images suitably to its target images. (b) changes the tops and bottoms attributes simultaneously.
The right-most columns are the results of multiplex face attribute transfer. The three rows from the bottom changes smiling female based on attributes of various hair colors and unsmiling female; and the top five rows show the results of the opposite experiment. (b) black hair (c) blond hair (d) brown hair. }
  \label{fig:multiplex}
  \end{center}
\end{figure}

\subsubsection{Multiplex Visual Attribute Transfer}

As we experimented the simple attribute transfer (changing one of many attributes) over the tasks, we ran the same experiment on multiplex attribute transfer where multiple attributes are changed simultaneously. While training models for multiplex attribute transfer, we randomly applied additional conversion: transferring one of non-target attribute randomly, in order to make the model robust to multiplex transfer. As \ref{fig:multiplex} shows, we could transfer a face with azimuth of $30 \degree$ into a car with azimuth of $-30 \degree$, change both top and bottom garments of a person, and transfer unsmiling blond hair woman to a smiling black hair woman, which means our experiment demonstrated that various attributes-based transfers could be successfully made.

\section{Conclusion}
This paper introduces an unsupervised visual attribute transfer using reconfigurable generative adversarial network. We demonstrate a single network can be used to transfer attributes among images in a controlled manner. In the future, unsupervised domain transfer can be explored to enable non-visual applications.

\renewcommand\refname{\subsubsection*{References}}
\small{
\bibliographystyle{unsrt}
\bibliography{ref}
}

\end{document}